\pdfoutput=1
\documentclass[]{collinear}

\title{\ycbench: Benchmarking AI Agents for Long-Term Planning and Consistent Execution }

\author{Muyu He*}
\author{Adit Jain*}
\author{Anand Kumar*}
\author{Vincent Tu*}
\author{Soumyadeep Bakshi}
\author{Sachin Patro}
\author{Nazneen Rajani}

\abstract{
As LLM agents tackle increasingly complex tasks, a critical question is whether they can maintain strategic coherence over long horizons: planning under uncertainty, learning from delayed feedback, and adapting when early mistakes compound. We introduce \ycbench, a benchmark that evaluates these capabilities by tasking an agent with running a simulated startup over a one-year horizon spanning hundreds of turns. The agent must manage employees, select task contracts, and maintain profitability in a partially observable environment where adversarial clients and growing payroll create compounding consequences for poor decisions. We evaluate 12 models, both proprietary and open-source, across 3 seeds each. Only three models consistently surpass the starting capital of \$200K, with Claude Opus 4.6 achieving the highest average final funds at \$1.27M, followed by GLM-5 at \$1.21M with 11$\times$ lower inference cost. Scratchpad usage, the sole mechanism for persisting information across context truncation, is the strongest predictor of success, and adversarial client detection is the primary failure mode, accounting for 47\% of bankruptcies. Our analysis reveals that frontier models still fail through distinct failure modes such as over-parallelization, demonstrating the capability gaps for  long-horizon performance. \ycbench is open-source, reproducible, and configurable at \url{https://github.com/collinear-ai/yc-bench}.
}
\correspondence{\url{research@collinear.ai}}
\date{\today}

\usepackage{amsfonts}
\usepackage{hyperref}
\usepackage{url}
\usepackage{booktabs}
\usepackage{xspace}
\usepackage{amsmath}
\usepackage{tabularx}
\usepackage{multirow}
\usepackage{subcaption}
\usepackage{enumitem}
\usepackage{titlesec}
\usepackage{listings}
\usepackage{xcolor}
\usepackage{float}
\usepackage[dvipsnames]{xcolor}
\usepackage{fvextra}
\usepackage{float}
\usepackage[T1]{fontenc}
\usepackage[utf8]{inputenc} 
\usepackage{tcolorbox}
\tcbuselibrary{breakable}
\usepackage{fvextra,times}
\usepackage{newunicodechar}

\lstdefinestyle{prompt}{
  basicstyle=\ttfamily\small,
  breaklines=true,
  frame=single,
  backgroundcolor=\color{gray!10},
  aboveskip=1em,
  belowskip=1em,
}

\titlespacing{\paragraph}{0pt}{0.5em}{0.5em}

\makeatletter
\def\blfootnote{\xdef\@thefnmark{}\@footnotetext}
\makeatother
\definecolor{ColOrange}{RGB}{242, 97, 37}

\newcommand{\ycbench}{\texttt{YC-Bench}\xspace}

\begin{document}

\maketitle




\blfootnote{\sffamily * indicates equal contribution}
\section{Introduction}

LLM-based agents have become increasingly capable in their ability to interact with environments and use tools. Because tasks may span 100+ tool calls, it's crucial for the agent to know how to plan well. As such, a growing body of work has focused on planning such as PlanBench \citep{valmeekam2023planbench} and BALROG \citep{paglieri2025balrog}. One key attribute that emerges in long-horizon planning is coherence:  over the course of hundreds or even thousands of interactions/steps, the agent must stay aligned with its goal, retain crucial facts or knowledge from its past, and avoid collapse into repetitive or hallucinated behavior. 

Existing works such as Vending-Bench (VB)~\citep{backlund2025vendingbench} and Vending-Bench 2, have aimed to test the ability of frontier models to maintain long-term coherence. These works have made meaningful progress in uncovering interesting planning failure modes and behaviors in frontier models, such as ``meltdown'' looping, hallucinations about non-existent facts, and even legal threats. They also have real-world randomness, including missed deliveries and unsatisfied customers. However, their environmental attributes have immediate tangible ramifications. For example, if the price is set incorrectly there, the agent observes a drop in sales the next day. In a real-world setting, running a business involves taking calculated risks without knowing the exact outcome immediately, and it involves taking actions that are suboptimal in the short term, in the hope of better long-horizon rewards. Further, if the current strategy is flawed and doesn't deliver maximum profits, the business must improve based on experience and adapt to a changing market landscape. 

\begin{figure}[!b]
    \centering
    \includegraphics[width=1\textwidth]{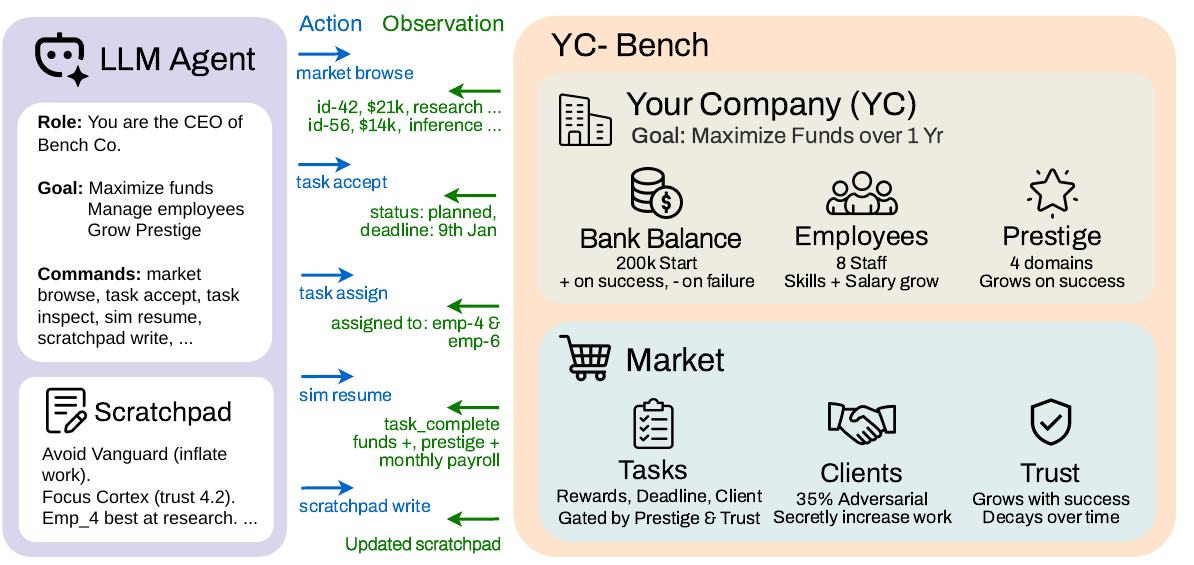}
    \caption{
Overview of \ycbench. The agent interacts with the environment through
   CLI commands (\textcolor{blue}{blue}) and receives structured observations (\textcolor{ForestGreen}{green}). The 
  environment tracks observable state (tasks, employees, finance,         
  prestige, client trust) and one hidden element: adversarial clients 
  whose work inflation must be inferred from repeated task failures.
  } 
    \label{fig:main_results}
\end{figure}

To fill this gap, we introduce \ycbench, a long-term coherence benchmark that evaluates an LLM agent's ability to run a simulated startup over a one-year horizon spanning hundreds of turns. The agent operates through a CLI tool interface, deciding each turn which contracts to accept from a marketplace, which employees to assign based on their skill profiles, and how to manage cash flow against recurring monthly payroll. Our environment is adversarial: roughly a third of clients are untrustworthy, and their tasks are designed to fail. The agent must learn to identify these clients by analyzing its own history of successes and failures, then avoid them going forward. This is another key distinction with VB, where the adversaries are known before commitment (e.g, adversarial suppliers). The benchmark tests long-term coherence through a 20-turn context window that forces the agent to use a persistent scratchpad for memory: agents that fail to record which clients are adversarial will repeat costly mistakes after their conversation history is truncated. Compounding dynamics reward sustained good decisions — trust builds with repeated clients, reducing future workload, while over-staffing tasks inflates payroll costs over time. Performance is measured by final company funds at year end, a single scalar that reflects the cumulative impact of hundreds of sequential decisions around task selection, resource allocation, and risk management. Our contributions can be summarized as follows {\color{blue}}:

\begin{enumerate}
    \item We introduce \ycbench, a POMDP-based benchmark with deterministic but unknown transition and observation dynamics. As illustrated in Figure~\ref{fig:main_results}, the LLM agent must plan and manage the day-to-day operations of a startup — accepting tasks from diverse clients and domains, and assigning them to employees with varying skill sets. \ycbench is designed to evaluate long-term planning and adaptive execution under delayed, sparse rewards.
\item In our main experimental results, we extensively test on frontier 
  models including GPT-5.4, GPT-5.4 Mini and Nano~\citep{openai2026gpt54},
   Claude Opus 4.6~\citep{anthropic2026opus46}, Claude Sonnet             
  4.6~\citep{anthropic2026sonnet46}, Gemini 3.1 Pro Preview, Gemini 3
  Flash, Gemini 3.1 Flash Lite~\citep{google2025gemini3}, Grok 4.20-beta  
  as well as popular open-source models like                      
  Qwen3.5-397B-A17B~\citep{qwen2026qwen35},
  GLM-5~\citep{glm5team2026glm5}, Kimi-K2.5~\citep{kimiteam2026k25}. Our main results are summarized in Section~\ref{sec:main_results} with Figure~\ref{fig:main_results} containing the averaged temporal runs.
  \item Through our error analysis of Section~\ref{sec:main_results} and Section~\ref{sec:error_analysis}, we observe that most model go bankrupt and there is still a significant gap in frontier models. Furthermore, we observe most models are unable to be profitable, tending to fall into adversarial client traps or under-staffing. There is also a significant gap in cost-efficiency with open-source models being more Pareto optimal than their counterparts.
\end{enumerate}

In summary, we introduce \ycbench, a long-term coherence benchmark that evaluates an agent's ability to simulate running a startup. Concretely, our benchmark tests an agent's ability to allocate resources in a complex organization under uncertainty, manage risk, and adapt to adversarial dynamics over long horizons. 

\section{Related Work}
\label{sec:related_work}

\ycbench sits at the intersection of several active research areas: LLM agent benchmarks, long-horizon planning, simulation-based evaluation, tool use and memory architectures.

\paragraph{LLM Agent Benchmarks. } 
The rapid growth of agentic capabilities of LLMs has spawned a rich ecosystem of benchmarks. AgentBench~\citep{liu2024agentbench} evaluates LLMs across eight distinct environments, including operating systems and games; and finds that poor long-term reasoning is the primary bottleneck for open-source models. SWE-bench~\citep{jimenez2024swebench} grounds evaluation in real GitHub issues, requiring agents to navigate codebases and generate correct patches; however, each task is an independent, bounded problem rather than a sustained sequential process.  GAIA~\citep{mialon2024gaia} tests general assistant capabilities across 466 real-world questions requiring tool use and multi-modal reasoning.  AgentBoard~\citep{ma2024agentboard} introduces the ``Progress Rate'' metric that captures incremental advancement rather than binary success, a design philosophy shared by \ycbench, which tracks revenue trajectories and prestige progression rather than pass/fail outcomes. TheAgentCompany~\citep{xu2024theagentcompany} embeds agents as digital workers in a simulated software company, finding that the best agent completes only 30\% of professional tasks. Both share \ycbench's emphasis on domain-grounded, consequential decision-making, but neither evaluates the sustained strategic coherence over hundreds of compounding turns that \ycbench requires.

\paragraph{Long-Horizon Planning and Reasoning. } 

Effective performance on \ycbench requires agents to maintain strategic coherence over hundreds of turns, a capability that pushes beyond the horizons tested by most existing work. PlanBench~\citep{valmeekam2023planbench} evaluates LLMs on classical planning domains (Blocksworld, Logistics), finding that even frontier models fall well short on plan generation and verification, providing evidence that apparent planning capabilities may be retrieval rather than genuine reasoning, a distinction that \ycbench's novel simulation setting, immune to pretraining contamination, is well-positioned to test. BALROG~\citep{paglieri2025balrog} is also closely related, evaluating LLMs across six procedurally generated game environments spanning easy to extremely challenging sequential tasks. Its key finding, that current frontier models still struggle significantly with longer-horizon tasks, directly motivates \ycbench's focus on sustained decision-making.

Most directly related to our work is Vending Bench (VB)~\citep{backlund2025vendingbench}, which served as a key inspiration for \ycbench. VB tests whether LLM agents can maintain coherent behavior while operating a simulated vending machine business, and finds that all models exhibit failure loops from which they rarely recover. \ycbench substantially extends this paradigm by introducing multi-domain task allocation, hidden employee skill rates that create an information-asymmetry puzzle, compounding financial dynamics (prestige decay, salary growth), and a multi-episode learning framework that tests whether agents can improve across restarts. \ycbench differentiates itself through its deterministic business-simulation setting and its status as an open-source, extensible, and configurable benchmark.

\paragraph{Simulation Environments and Strategic Games. }

\ycbench belongs to a growing family of simulation-based benchmarks that test AI decision-making in complex, multi-turn environments. Generative Agents~\citep{park2023generative} introduced LLM-powered agents that simulate believable human behavior in a sandbox, using memory storage, reflection, and dynamic retrieval for planning; their architecture for persistent agent memory directly informs \ycbench's scratchpad-based memory carryover between episodes. Cicero~\citep{meta2022cicero} demonstrated that LLMs can handle negotiation and long-term planning under imperfect information in Diplomacy, capabilities analogous to those tested in \ycbench's information-asymmetric setting where agents must infer hidden employee skill rates from task outcomes. CivRealm~\citep{qi2024civrealm} provides a Civilization-based benchmark requiring resource management and diplomacy over extended horizons, but both RL- and LLM-based agents struggle with the full game's complexity, motivating benchmarks like \ycbench that isolate specific aspects of strategic decision-making in a controlled, deterministic setting.

\begin{figure}[!b]
    \centering
    \includegraphics[width=0.92\textwidth]{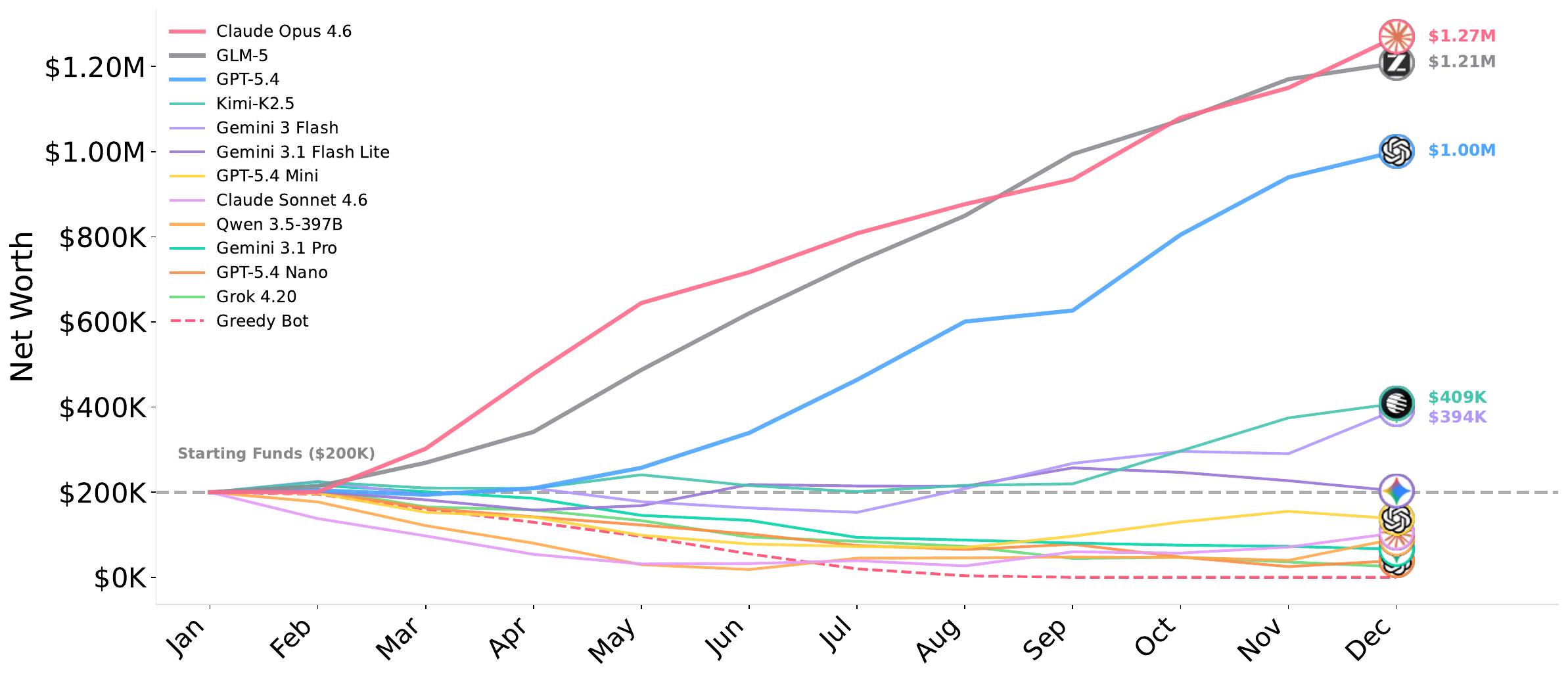}
    \caption{ Out of the $12$ models that we benchmark on \ycbench, $5$ models are profitable and  only $3$ turn a substantial profit ($5\times$ profit). The figure plots the funds across time averaged across three seeds for each model.  Comprehensive results can be found in Appendix~\ref{app:main_result_stats}. 
    }
    \label{fig:main_results_image}
\end{figure}

\section{YC-Bench}
This section describes the \ycbench benchmark, which we mathematically formalize as a Partially Observable Markov Decision Process (POMDP) and describe the different components and dynamics involved in the benchmark including Tasks, Employees, Clients, Memory and Simulation Clock. 

\paragraph{Environment Definition. }
We formalize \ycbench as a Partially Observable Markov Decision Process (POMDP) $\mathcal{M} = (\mathcal{S}, \mathcal{O}, \mathcal{A}, \mathcal{T}, R)$, where $\mathcal{S}$ is the state space, $\mathcal{O}$ is the observation space, $\mathcal{A}$ is the action space, $\mathcal{T}: \mathcal{S} \times \mathcal{A} \rightarrow \mathcal{S}$ is the deterministic transition function, and $R: \mathcal{S} \times \mathcal{A} \rightarrow \mathbb{R}$ is the reward function defined as the net change in company funds. The agent operates a company over a one-year simulated horizon, starting with \$200{,}000 in funds. At each turn $t$, the agent receives a partial observation $o_t \in \mathcal{O}$ and selects a sequence of actions $\mathbf{a}_t \in \mathcal{A}^*$, after which the environment transitions deterministically to $s_{t+1} = \mathcal{T}(s_t, \mathbf{a}_t)$. The agent controls time progression explicitly by issuing a \texttt{sim resume} command; between time advances, it may issue arbitrarily many actions within a single turn. The episode terminates when funds drop below zero (bankruptcy) or the horizon ends. The objective is to maximize final funds $f_T$.

We now describe the core mechanics of the environment that determine agent performance. At a high level, the agent must balance competing pressures: completing tasks to generate revenue while avoiding adversarial clients that inflate work requirements, building domain prestige to unlock higher-reward tasks, and managing a payroll that grows with every successful completion. The full agent action space is in Appendix~\ref{app:action-space}.

\textbf{Tasks and domains.} The agent's sole source of income is completing tasks accepted from a marketplace. Each task belongs to one of four industry \emph{domains}---training, inference, research, and data engineering---and is issued by a client. A task comes with a predetermined reward, a deadline that activates upon acceptance, and a \emph{work quantity} in that domain that employees must complete before the deadline. Accepting a task may require a minimum \emph{prestige} level in its domain and a minimum \emph{trust} level with its client. The agent maintains a prestige level in each domain: higher prestige unlocks higher-reward tasks and linearly scales their payout. Completing a task grants both funds and a prestige increase in its domain, creating a progression dynamic where early task selection determines which high-value tasks become accessible later. If the agent fails to complete a task by its deadline, it incurs a penalty of 35\% of the advertised reward and a prestige reduction, incentivizing accurate estimation of completion time relative to employee capacity.

\textbf{Employees and productivity.} The company has a fixed roster of staff members that cannot be hired or fired. Each employee has a per-domain \emph{productivity} level, the quantity of work they complete per hour in that domain. The agent can query productivity levels directly via \texttt{employee list}. Each employee also has a tier label (junior, mid, or senior) reflecting their average productivity, but productivity distributions are spiky: a senior-tier employee may have senior-level throughput in training but junior-level throughput in research. Effective employee assignment therefore requires the agent to match employees to tasks based on their domain-specific strengths rather than relying on tier alone. Upon successful task completion, assigned employees receive a productivity boost, a percentage increase to their productivity in the task's domain, capped at a maximum rate. This makes employees more productive over time, rewarding consistent domain-specialized assignment. However, each completion also triggers a \emph{salary bump}, a fixed raise based on the employee's tier. The monthly payroll therefore grows monotonically with the number of completed tasks, creating pressure to prioritize high-reward tasks that outpace rising costs.

\textbf{Clients.} Each task is issued by one of several clients. The agent builds \emph{trust} with a client by completing its tasks successfully; higher trust reduces the work required on future tasks from that client and unlocks higher-tier tasks. However, completing a task for one client slightly decays trust with all others, so the agent must choose which client relationships to invest in.

\textbf{Adversarial clients.} A subset of clients are \emph{adversarial}: after the agent accepts one of their tasks, the environment inflates the work quantity, making the deadline nearly impossible to meet. Adversarial status is hidden and never revealed directly. Crucially, adversarial clients offer competitively high rewards, so the agent cannot simply filter by price. The agent must instead infer which clients are adversarial from the pattern of repeated task failures.

\textbf{Observations.} At the start of each turn, the agent receives a structured status summary containing the current timestamp, funds, monthly payroll, runway estimate, active task count, and a list of events since the previous turn (task completions, fund changes, salary bumps, deadline margins). All other state information (employee skill tiers, market tasks, client trust levels, per-task progress) must be actively queried through observe actions. The key hidden quantity is client reliability levels, which must be inferred from the pattern of task outcomes.

\textbf{Simulation Clock.} The agent controls the passage of simulated time explicitly. Calling \texttt{sim resume} advances the clock to the next scheduled event, typically a task checkpoint (25\%, 50\%, 75\%, or 100\% completion), a monthly payroll deduction, or the horizon end. Between events, the agent may issue arbitrarily many actions to reassess strategy, accept new tasks, or reassign employees. Work progresses only during business hours (weekdays), and payroll is deducted on the first business day of each month.

\textbf{Memory.} The agent's conversation history is truncated to the most recent $K$ turns (we use $K{=}20$ in our experiments). To maintain long-term coherence, the agent may write to a persistent \texttt{scratchpad} that is injected into the system prompt on every turn. The scratchpad is the agent's sole mechanism for retaining information across context truncation, for example recording which clients are unreliable, which employees are strongest in which domains, or strategic rules derived from past performance. This design does not bias toward any particular memory strategy; instead, it tests whether agents can autonomously determine what information is worth persisting.

\begin{figure}[!t]
  \centering
  \begin{subfigure}[t]{0.48\textwidth}
    \centering
    \includegraphics[width=0.9\textwidth]{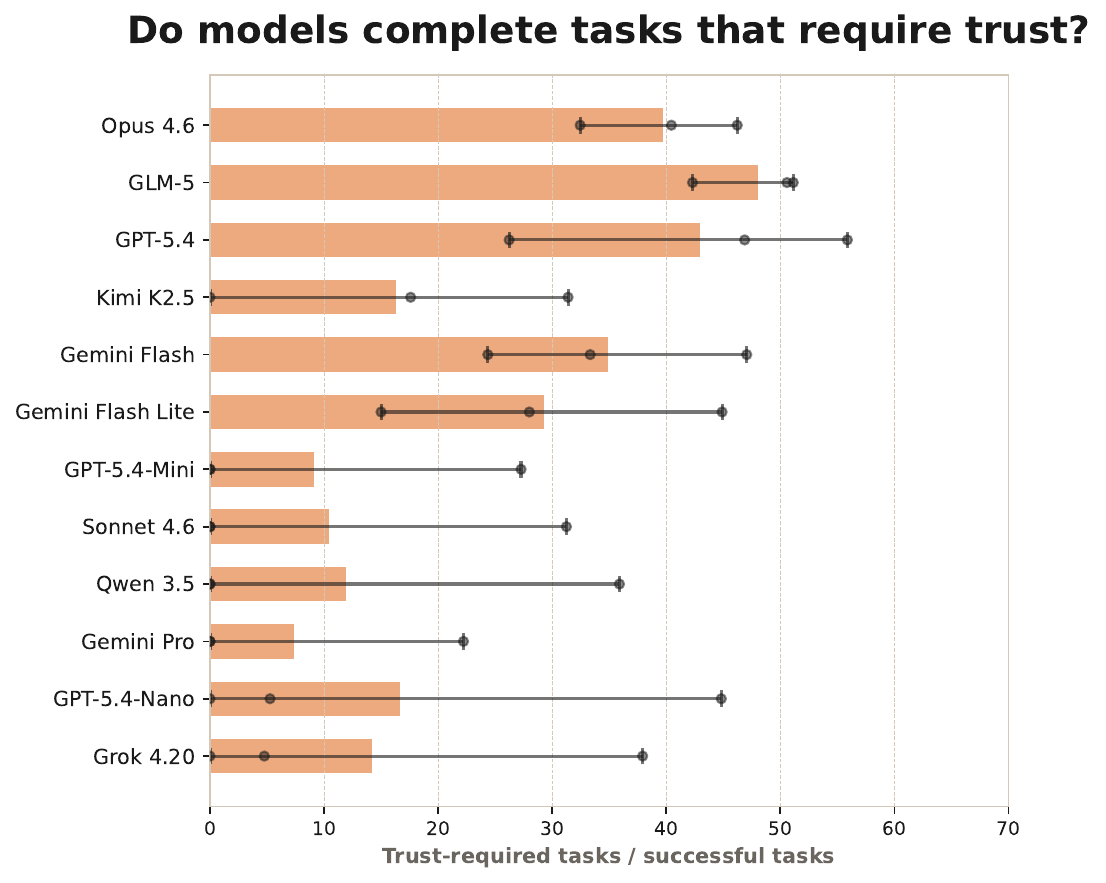}
    \caption{Proportion of successfully completed tasks that require client trust to complete. Error bars show per-seed range across 3 seeds.}
    \label{fig:trust_combined_a}
  \end{subfigure}
  \hfill
  \begin{subfigure}[t]{0.48\textwidth}
    \centering
    \includegraphics[width=0.86\textwidth]{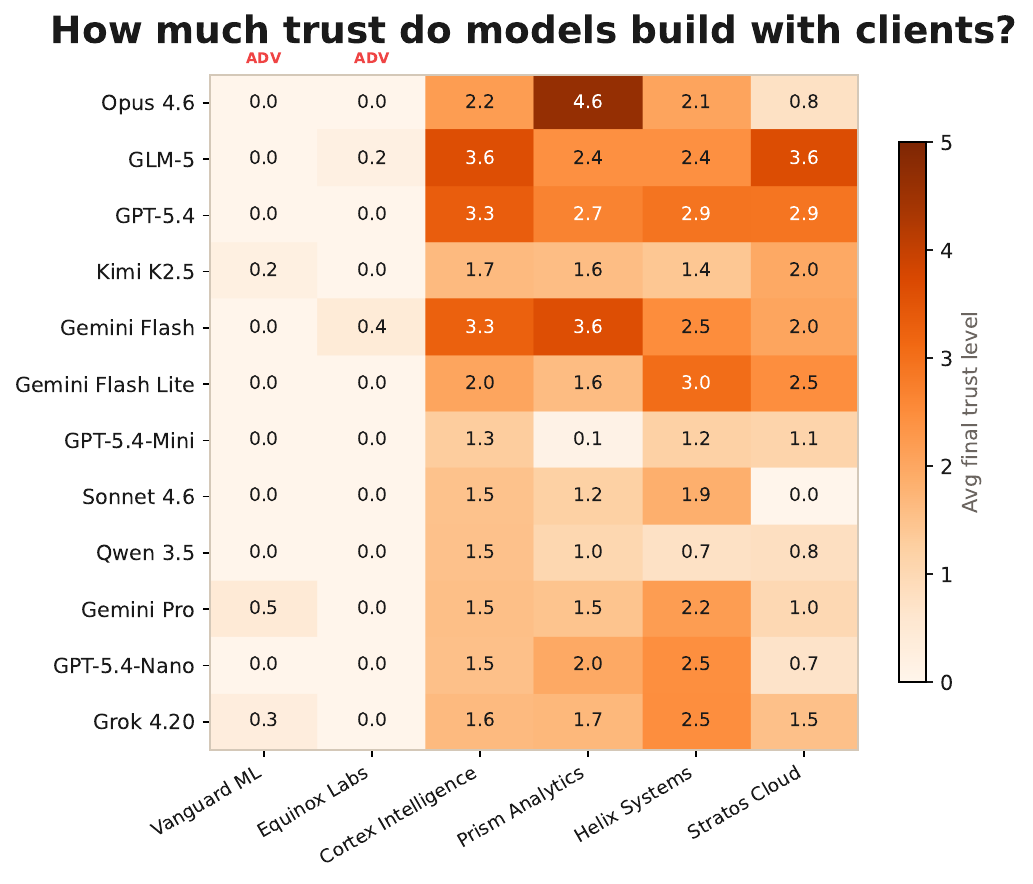}
    \caption{Final trust level for different models per client averaged across seeds (ADV=Adversarial).}
    \label{fig:trust_combined_b}
  \end{subfigure}
\caption{We observe that better models are able to build client trust over time by strategically selecting clients. What is surprising is smaller distilled models (Sonnet-4.6) do worse than contemporaries (Gemini-3-Flash) unlike VB.}
  \label{fig:trust_combined}
\end{figure}
\section{Experiments}

\subsection{Experimental Setup}

\paragraph{Models} We evaluate 12 models spanning seven providers:
GPT-5.4, GPT-5.4-Mini, and GPT-5.4-Nano (OpenAI); Claude Opus 4.6 and
Claude Sonnet 4.6 (Anthropic); Gemini 3.1 Pro, Gemini 3 Flash, and
Gemini 3.1 Flash Lite (Google); Qwen 3.5-397B (Alibaba); GLM-5 (Zhipu
AI); Kimi-K2.5 (Moonshot AI); and Grok 4.20 (xAI). Each model is
evaluated on 3 seeds using the \texttt{default} preset, for a total of
36 LLM runs. We use the \texttt{LiteLLM} framework for inference and
OpenRouter as the model provider. Full configuration hyperparameters are
listed in Appendix~\ref{app:config} and the agent system prompt can be found in Appendix~\ref{app:sys-prompt}.
   
\paragraph{Baseline} We compare against a greedy baseline that, in each turn, accepts the highest-reward task available on the market, assigns all employees to it, and advances the simulation clock. This baseline does not check client history, specialize employees by domain, nor does it use the scratchpad. An LLM-based agent should be able to surpass the greedy baseline by intelligently selecting the right tasks from non-adversarial clients and the correct employees for that task over the long-term.

\subsection{Main Results}\label{sec:main_results}

\paragraph{\ycbench reveals large performance gaps between frontier models that score similarly on standard benchmarks.}
As shown in Figure~\ref{fig:main_results}, of the twelve models evaluated, only three (Claude Opus 4.6, GLM-5, and GPT-5.4) exceed \$1{,}000{,}000 in average final funds, 2--3$\times$ higher than the next-best model. Only five models turn a profit on average; the remaining seven finish below their \$200{,}000 starting capital, and several go bankrupt in at least one seed.
Figure~\ref{fig:main_results} shows that the divergence emerges by February--March, roughly 60 days into the simulation. Top models concentrate on one or two clients early, triggering a trust snowball: each success reduces future task workloads (up to 50\%), enabling more completions per month, which builds further trust. Models that spread work across many clients never reach meaningful work reduction and enter a payroll-driven decline. 

\begin{figure}[!b]
  \centering
  \begin{subfigure}[t]{0.48\textwidth}
    \centering
    \includegraphics[width=\textwidth]{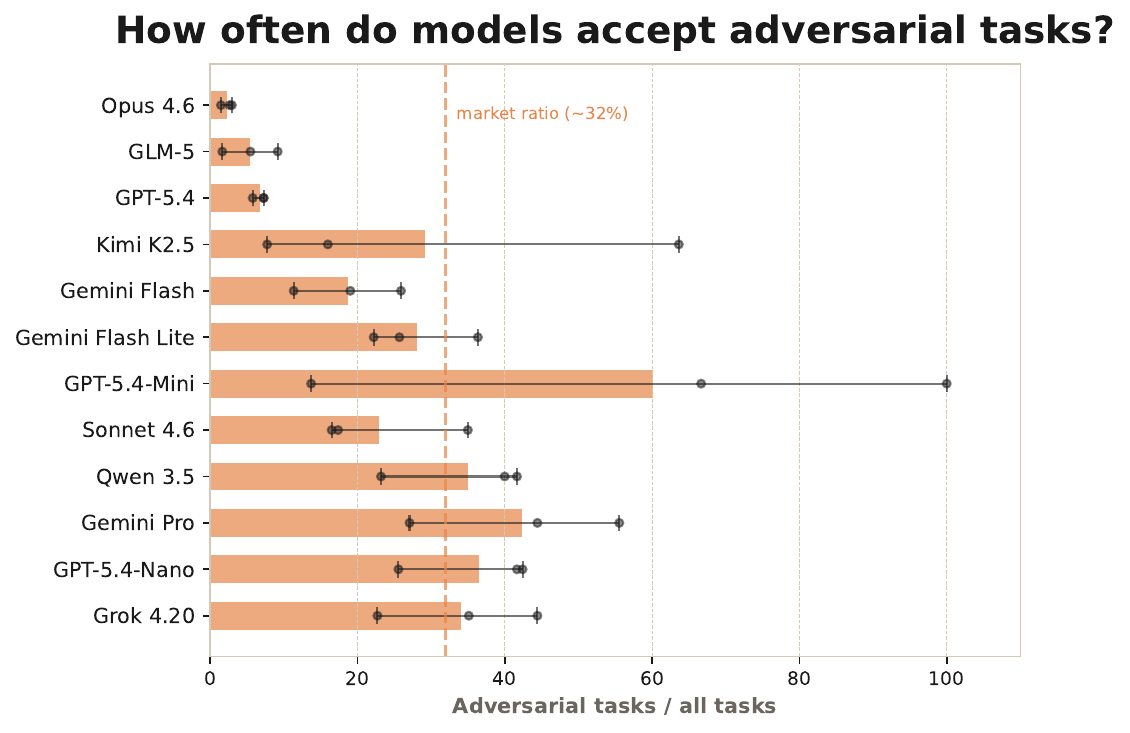}
    \caption{Average ratio of adversarial client tasks among all accepted tasks. Error bars show per-seed range. Dashed line indicates the natural market share of adversarial tasks (~32\%).}
    \label{fig:adversarial_combined_a}
  \end{subfigure}
  \hfill
  \begin{subfigure}[t]{0.48\textwidth}
    \centering
    \includegraphics[width=0.9\textwidth]{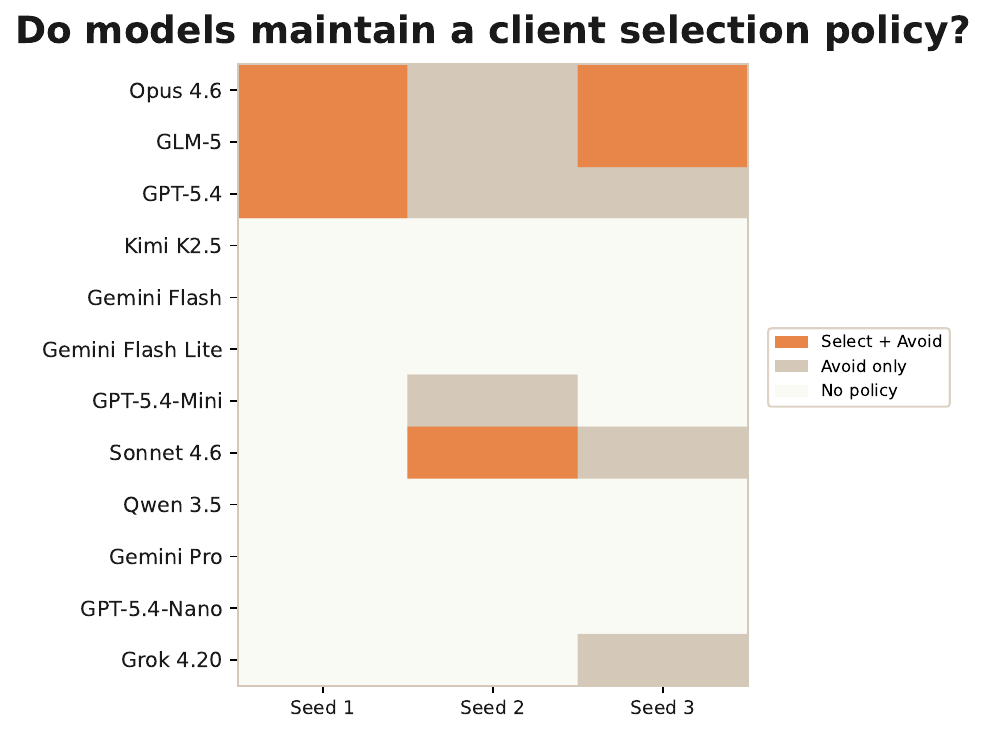}
    \caption{Client selection policy as observed in agent scratchpads for each of the $3$ seeds.}
    \label{fig:adversarial_combined_b}
  \end{subfigure}
  \caption{Analyzing how the models deal with adversarial clients, who have appealing rewards when accepting a task but have a lot more work than claimed (scope creep).}
  \label{fig:adversarial_combined}
\end{figure}

\paragraph{Only a few models can stick to improving client trust in the long term; most models choose clients indiscriminately.} Figure~\ref{fig:trust_combined_a} shows the ratio of completed tasks that require the model to gain non-zero trust level with the client that issues the task. 
A model is motivated to choose tasks that require trust if they are accessible because they come with both higher rewards and smaller workloads.
Despite the clear motivation to selectively focus on a particular subset of clients, most models do not build trust consistently with clients.
This is seen in Figure~\ref{fig:trust_combined_a}, where we see most models maintain a minimal trust level (1-2) with most clients, barring themselves from access to tasks with higher returns.
Analysis of the model scratchpads in Figure~\ref{fig:adversarial_combined_b} shows that only 4 out of 10 models across 6 out of 30 runs explicitly maintain a "whitelist" of clients to work with. 
The rest of the runs distribute tasks among clients discriminately: 'Avoid only' shows that the model only maintains a list of clients not to work with; 'Select + Avoid' shows that the model in addition maintains a list of clients to focus on; 'No policy' means no list is maintained.

\paragraph{Identifying adversarial clients in the market remains a challenge for all but a few models.}  Figure~\ref{fig:adversarial_combined_a} computes the average ratio of adversarial client tasks in all tasks accepted by each model across seeds. 
Half of all the models accept adversarial tasks at a rate higher than the natural market share of those tasks, showing that these models either are indifferent to the adversarial clients or come to prefer them over good clients by misjudgment. 
As shown in Figure~\ref{fig:adversarial_combined_b}, two-thirds of all runs make no mention of ``blacklisting'' any adversarial client in their strategy, leading them to choose clients indiscriminately.

However, as Figure~\ref{fig:adversarial_combined_a} also reveals, the three top-performing models on average strategically accept adversarial tasks significantly less frequently than the baseline, at 1/4 the rate of the next best model. 
As the scratchpad reveals, the top-performing models correctly spot the increase in work quantity before and after accepting the problematic tasks and subsequently write explicit guidelines to never accept tasks from those clients. Despite occasional inconsistencies in following the guidelines, these instructions effectively prevent excessive task failures caused by these adversarial tasks.
\begin{figure}[!t]
  \centering
  \begin{subfigure}[t]{0.48\textwidth}
    \centering
    \includegraphics[width=0.90\textwidth]{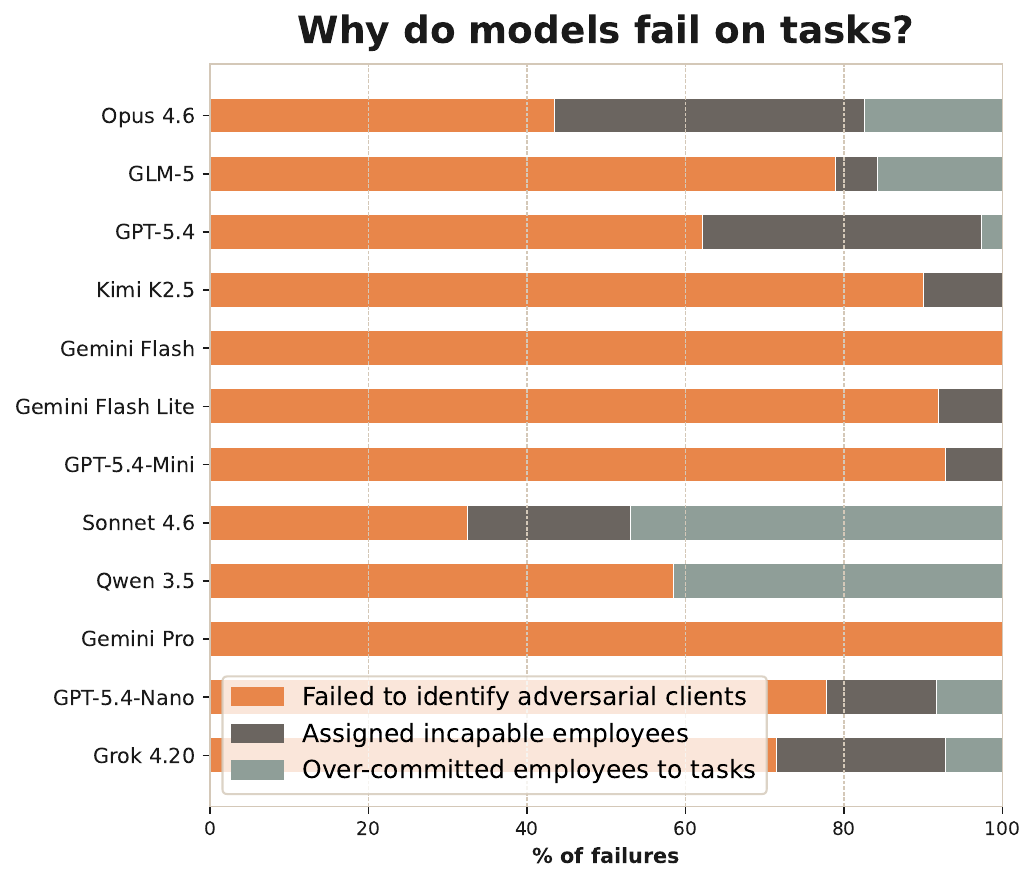}
    \caption{Failure mode breakdown. Models miss task deadlines due to (1) failing to identify adversarial clients; (2) assigning incapable employees; or (3) over-committed employees to tasks.}
    \label{fig:failure_and_cost_a}
  \end{subfigure}
  \hfill
  \begin{subfigure}[t]{0.48\textwidth}
    \centering
    \includegraphics[width=0.9\textwidth]{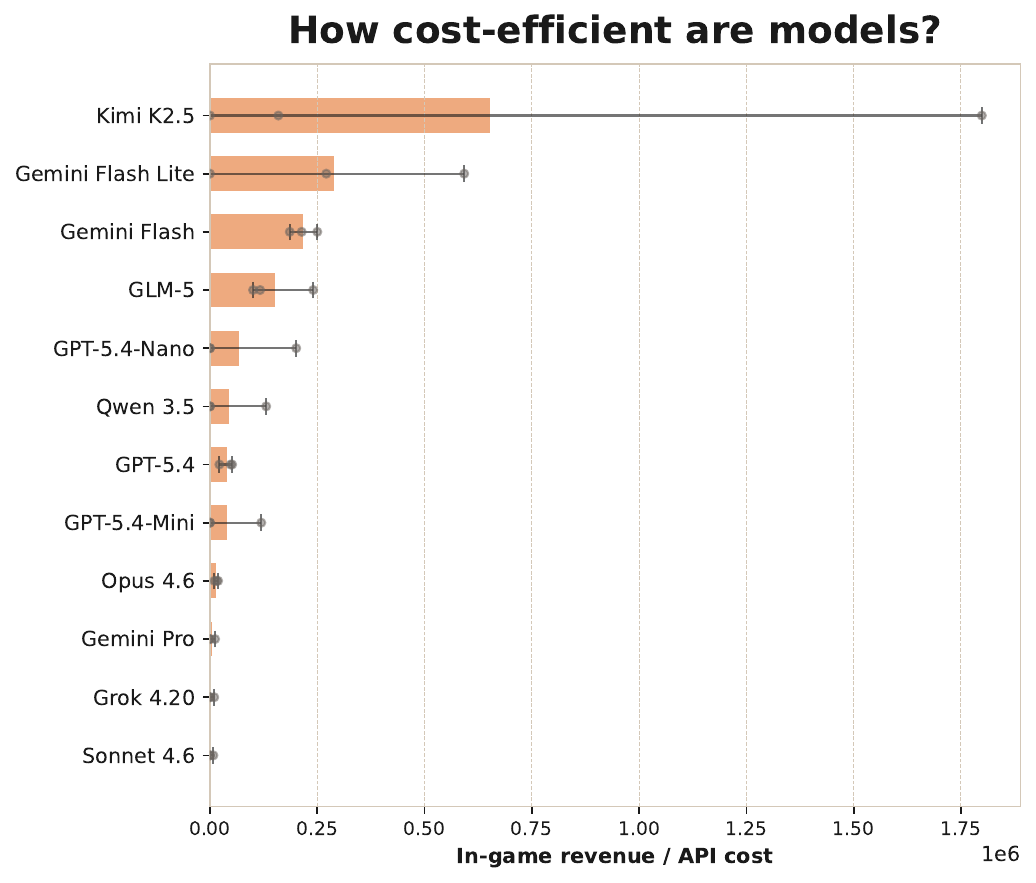}
    \caption{ Cost efficiency measured as revenue per dollar of API cost. Error bars show per-seed range. }
    \label{fig:failure_and_cost_b}
  \end{subfigure}
  \caption{Other than identifying adversarial clients there are two other primary task failure modes, especially wrong employee assignment. We observe that Kimi-K.5 is the most cost efficient, whereas the second highest ranking model, GLM-5 is substantially ($10
  \times$) better than Opus-4.6 on cost efficiency. }
  \label{fig:failure_and_cost}
\end{figure}

\paragraph{Even the top-performing models still suffer from repercussions of suboptimal employee assignment decisions.}

Figure~\ref{fig:failure_and_cost_a} breaks down all failed tasks for each model across runs by the cause of failure. 
Although failure to identify adversarial clients is still the most common pitfall by design, 7 out of 11 models have a substantial portion of failure attributed to suboptimal employee assignment.
Specifically, they either assign employees whose productivity can never theoretically complete the task in time or spread employees' effort across multiple tasks so that their throughput on each task cannot meet the minimal requirement.
When the model assigns employees to tasks, they have perfect information about the amount of work in each task and the employee's productivity relevant to its domains.
As a result, the failure stems from not properly estimating the speed of completion against the deadline and not taking into account the fact that working on multiple tasks slows each employee down.

\paragraph{Models drastically differ in cost-efficiency, with medium-performing models such as Kimi-K2.5 striking a superior balance between cost and revenue over top-performing ones.}
Figure~\ref{fig:failure_and_cost_b} calculates how much in-game revenue in million dollars each model generate per dollar in API costs. 
\texttt{Kimi-K2.5} achieves superior performance over all other models, having a $2.5\times$ performance gap with the next most cost-effective model, \texttt{Gemini-3-Flash}.  
The three top-performing models significantly lag behind in terms of cost-effectiveness, especially the strongest model \texttt{Claude-Opus-4.6}, despite yielding $3\times$ the return of the most cost-effective model.
The ranking has practical weight for production because it shows how much actual return is to be expected from each model if we take the inference cost into account. Refer to Appendix~\ref{app:summary_token_use} for summary statistics on token usage, time, and cost.

\section{Error Analysis}\label{sec:error_analysis}

We analyze four models to understand the behavioral mechanisms behind their performance, focusing on their error modes. The brackets are (Final Revenue, Number of Bankruptcies).

\paragraph{Claude Opus 4.6 (\$1.27M avg, 0/3 bankrupt).}
Opus actively uses the scratchpad to strategize, rewriting it $\sim$34 times per run across four topics: calibrating environment mechanics from observed completions, following a one-task-at-a-time workflow, building a client blacklist, and tracking per-client success rates to optimize trust-gated task selection. These topics emerge in a consistent order: calibration dominates the first $\sim$10 turns, workflow rules solidify by Turn $\sim$20, adversarial clients are flagged by Turn $\sim$25, and trust-based optimization continues for the remainder of the run. It also inspects every task before acceptance ($\sim$155 calls per run). Despite this, Opus is not flawless: in Seed 2 (\$750K), it accepts a task from a blacklisted client in August and accumulates 9 legitimate failures in the second half as higher-prestige tasks prove harder to complete.

\paragraph{Gemini 3 Flash (\$394K avg, 0/3 bankrupt).}
Flash averages 2 scratchpad entries per run, rarely inspects tasks, and executes an identical 4-command cycle every turn: accept, assign all 8 employees, dispatch, resume. This lack of adaptation is costly: Flash accepts 12 adversarial tasks across the run, all of which fail, capping its earnings well below the top tier. Its 0/3 bankruptcy rate reflects sufficient throughput to absorb these losses, not the ability to avoid them.

\paragraph{Claude Sonnet 4.6 (\$103K avg, 2/3 bankrupt).}
Sonnet exhibits a \emph{reasoning--execution gap}: it derives correct strategies but fails to act on them. At Turn 7 in Seed 1, it writes a correct feasibility formula to the scratchpad (``required\_qty / total\_rate must be $<$ deadline hours'') and a ``one task at a time'' rule. It then ignores both: Turn 8 accepts four tasks without inspection, and over the full run it averages 7.23 concurrent active tasks (max 16). Of its 17 failed tasks, 41\% result from understaffing (tasks that would succeed with all 8 employees assigned), 18\% from assigning employees poorly matched to the task domain, and 12\% from re-accepting adversarial clients without ever recording the pattern in its scratchpad. Sonnet averages 8 scratchpad entries per run but stops updating early, leaving stale financial data (\$152K) while the actual balance drops to $-$\$3K.

\paragraph{Grok 4.20 (\$14K avg, 2/3 bankrupt).}
Grok shows \emph{aware inaction}: its scratchpad accurately identifies critical issues (``Runway down to 1 month,'' ``Avoid Equinox'') but these observations do not translate into changed behavior: it accepts a task from Vanguard ML (0\% historical success rate) with 6 days of runway remaining. It also does not manage its active task portfolio, leaving one task accepted in March uncompleted and uncancelled for 81 days until bankruptcy. Grok averages 0.92 commands per turn, the lowest of any model evaluated (Table~\ref{tab:behavioral_stats}), suggesting it spends most turns on observation and deliberation rather than task execution.

Overall, the four failure profiles reveal a spectrum of long-term incoherence. Flash fails entirely due to the absence of reflection. Grok fails despite accurate reflection, unable to close the loop between diagnosis and action. Sonnet fails from reflection that is structurally correct but temporally inconsistent, ie, rules written and immediately abandoned. Only Opus achieves sustained, self-correcting reflection, though even it is not immune to occasional blacklist violations. This spectrum suggests that long-horizon coherence is not a single capability but a pipeline: perceive → record → retrieve → act consistently, and current models fail at different stages of that pipeline.

\begin{figure}[!t]
    \centering
    \includegraphics[width=\textwidth]{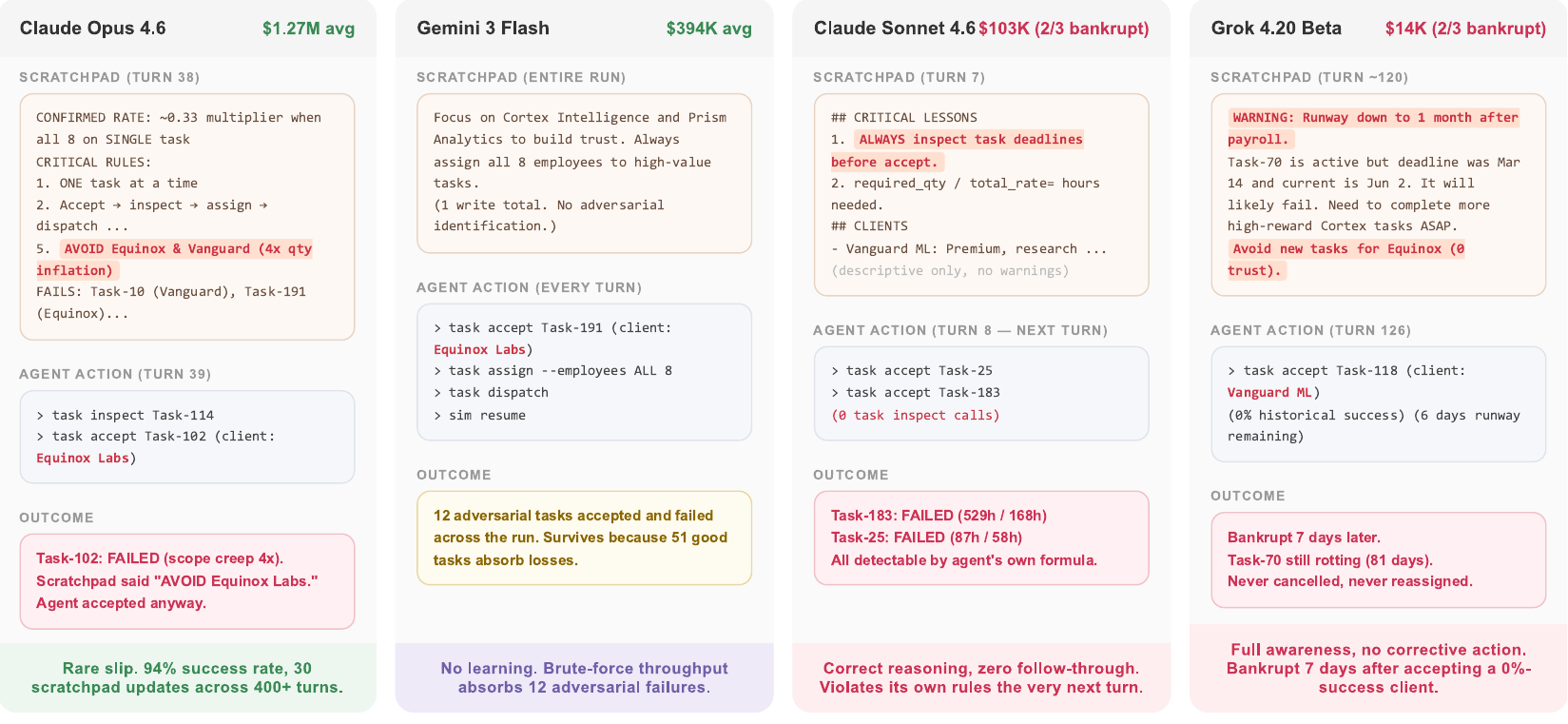}
    \caption{
Representative failure moments for four models. Each panel shows the  
  scratchpad state, agent action, and outcome. From left to right: Opus   
  violates its own blacklist; Flash blindly accepts adversarial tasks     
  through a rigid loop; Sonnet writes a correct rule and immediately      
  ignores it; Grok diagnoses its bankruptcy but takes no corrective       
  action.
  }
\end{figure}

\section{Conclusion}

In this work, we present \ycbench, a simulation-based benchmark that comprehensively evaluates LLMs' long-term coherence, planning, and consistent execution abilities. The benchmark introduces design components that can only be tackled by a model excelling in long-term optimization: evolving client trust relationships, disruption by adversarial clients, and compounding financial pressure from company growth.

\ycbench reveals that long-horizon coherence remains a critical, under-tested capability: only 3 of 12 frontier models grew their starting capital, and scratchpad usage was the strongest single predictor of success. Adversarial client detection, accounting for $47\%$ of bankruptcies, exposed a failure mode that benchmarks with immediate environmental feedback cannot surface. Our error analysis reveals a recurring reasoning–execution gap as they derive correct strategies but consistently fail to act on them, suggesting that deliberation and execution are not yet unified capabilities in current frontier models.

\textbf{Limitations and Future Work. }
The current environment holds several simplifying assumptions worth relaxing in future iterations: employees cannot be hired or fired, disruptions are limited to adversarial clients rather than random exogenous events, and all quantitative signals (work quantities, productivity rates) are provided numerically rather than in natural language. Introducing these elements would create a richer, more realistic decision surface and likely surface additional failure modes in frontier models. We will release \ycbench as an open-source, configurable benchmark and invite the community to stress-test future models against its compounding, adversarial dynamics.

\bibliography{colm2026_conference}
\bibliographystyle{iclr2026_conference}

\appendix
\clearpage
\appendix
\section*{Appendix}

\section{Action Space}
\label{app:action-space}

The complete list of CLI commands the model can use in \ycbench is provided in Table~\ref{tab:actions}.
Together they define the possible action space for the model.

\begin{table}[ht]
\centering
\small
\begin{tabularx}{\textwidth}{@{}llX@{}}
\toprule
\textbf{Category} & \textbf{Command} & \textbf{Effect} \\
\midrule
Observe & \texttt{company status} & Funds, prestige, payroll \\
Observe & \texttt{employee list} & Names, tiers, salaries, productivity \\
Observe & \texttt{market browse} & Available tasks with client, reward, domains \\
Observe & \texttt{task list} & Accepted tasks with status and progress \\
Observe & \texttt{task inspect --task-id T} & Per-domain progress, deadline, assignments \\
Observe & \texttt{client list} & Client trust levels and tiers \\
Observe & \texttt{client history} & Per-client success/failure counts \\
Observe & \texttt{finance ledger} & Full transaction history \\
\midrule
Task & \texttt{task accept --task-id T} & Accept from market; starts deadline \\
Task & \texttt{task assign --task-id T --employees E} & Assign employees to task \\
Task & \texttt{task dispatch --task-id T} & Begin work on assigned task \\
Task & \texttt{task cancel --task-id T --reason R} & Abandon task; prestige penalty \\
\midrule
Sim & \texttt{sim resume} & Advance clock to next event \\
\midrule
Memory & \texttt{scratchpad write --content C} & Overwrite persistent notes \\
Memory & \texttt{scratchpad append --content C} & Append to persistent notes \\
\bottomrule
\end{tabularx}
\caption{Agent action space. All actions are CLI commands with arguments constructed as free-form text. Multiple actions may be issued per turn.}
\label{tab:actions}
\end{table}

\section{Default Configuration for \ycbench}
\label{app:config}

The default configuration parameter for each variable in the environment is provided in Table~\ref{tab:default_config}.

\begin{table}
\centering
\footnotesize
\begin{tabularx}{\columnwidth}{@{}l l r >{\raggedright\arraybackslash}X@{}}
\toprule
\textbf{Category} & \textbf{Parameter} & \textbf{Value} & \textbf{Description} \\
\midrule
Simulation & Horizon & 1 yr & Evaluation period \\
Simulation & Auto-advance & 5 turns & Idle turns before forced advance \\
Simulation & Business hours & 9--18 & Weekdays only \\
\midrule
Workforce & Employees & 8 & Fixed roster \\
Workforce & Initial funds & \$200K & Starting capital \\
Workforce & Salary bump & 1\%/task & All assigned employees \\
\midrule
Market & Market tasks & 200 & Initial pool \\
Market & Browse limit & 50 & Visible per query \\
Market & Clients & 6 & Total count \\
\midrule
Prestige & Range & [1, 10] & Per domain \\
Prestige & Decay & 0/day & No passive loss \\
Prestige & Reward scale & 0.30 & Reward multiplier \\
Prestige & Req.\ prestige & Tri(1,5,1) & Task requirement dist. \\
\midrule
Deadlines & Qty/day & 150 & For deadline calc \\
Deadlines & Min days & 7 & Minimum deadline \\
Deadlines & Fail penalty & 35\% & Of advertised reward \\
Deadlines & Cancel penalty & $1.5\times\delta$ & Prestige lost \\
\midrule
Trust & Max & 5.0 & Per client ceiling \\
Trust & Build rate & 5.0 & $\sim$5 tasks to significant \\
Trust & Work reduction & 50\% & At max trust \\
Trust & Gated fraction & 30\% & Tasks requiring trust \\
Trust & Focus pressure & 0.3 & Cross-client decay \\
\midrule
Adversarial & Fraction & 35\% & Adversarial clients \\
Adversarial & Scope creep & $\geq 3.0\times$ & Work inflation floor \\
\midrule
Salary Tiers & Junior (50\%) & \$2--4K/mo & Rate 1--4 units/hr \\
Salary Tiers & Mid (35\%) & \$6--8K/mo & Rate 4--7 units/hr \\
Salary Tiers & Senior (15\%) & \$10--15K/mo & Rate 7--10 units/hr \\
\midrule
Distributions & Task reward & Tri(\$2K,\$12K,\$5K) & Per task \\
Distributions & Domain count & 1 & Domains per task \\
Distributions & Work qty & Tri(400,1500,800) & Units per domain \\
\midrule
Memory & Context window & 20 turns & Before truncation \\
Memory & Scratchpad & Persistent & In system prompt \\
\bottomrule
\end{tabularx}
\caption{Default configuration hyperparameters for \ycbench.}
\label{tab:default_config}
\end{table}

We selected this configuration because it is adequately difficult without being unfair or unnecessarily difficult for frontier models. 

\section{System Prompt}
\label{app:sys-prompt}

The system prompt for each evaluation run is provided as follows. The model can access the system prompt in the chat history at all times during each turn.
\lstset {
    captionpos=b,    
    frame=tb,        
}
\begin{lstlisting}[style=prompt,caption={Agent system prompt.},numbers=none]
You are the CEO of a startup in a business simulation. Maximize funds and prestige while avoiding bankruptcy.

All actions use `yc-bench` CLI commands via `run_command`. All return JSON.

## Core Workflow (repeat every turn)

**You must always have active tasks running. Every turn, follow this loop:**

1.`yc-bench market browse` - pick a task
2. `yc-bench task accept --task-id Task-42` - accept it
3. `yc-bench task assign --task-id Task-42 --employees Emp_1,Emp_4,Emp_7` - assign employees (check `employee list` for skill rates)
4. `yc-bench task dispatch --task-id Task-42` - start work
5. `yc-bench sim resume` - advance to next event (requires active tasks)

Run multiple tasks concurrently when possible. Accept -> assign -> dispatch a second task before calling sim resume.

**Use `yc-bench scratchpad write`** to save strategy notes - your conversation history is truncated after 20 turns, but scratchpad persists in the system prompt. Write reusable rules, not one-off observations.

## Commands

### Observe
- `yc-bench company status` - funds, prestige, payroll
- `yc-bench employee list` - employees with skill rates per domain
- `yc-bench market browse [--domain X] [--reward-min-cents N] [--limit N]` - available tasks
- `yc-bench task list [--status X]` - your tasks
- `yc-bench task inspect --task-id Task-42` - task details
- `yc-bench client list` - clients with trust levels
- `yc-bench client history` - per-client success/failure rates
- `yc-bench finance ledger` - financial history

### Act
- `yc-bench task accept --task-id Task-42` - accept from market
- `yc-bench task assign --task-id Task-42 --employees Emp_1,Emp_4,Emp_7` - assign employees (comma-separated)
- `yc-bench task dispatch --task-id Task-42` - start work (must assign first)
- `yc-bench task cancel --task-id Task-42 --reason "text"` - cancel (prestige penalty)
- `yc-bench sim resume` - advance time
- `yc-bench scratchpad write --content "text"` - save notes
- `yc-bench scratchpad append --content "text"` - append notes

## Key Mechanics

- **Salary bumps**: completed tasks raise salary for every assigned employee. More employees assigned = higher payroll growth.
- **Throughput split**: employees on multiple active tasks split their rate (rate/N). Two tasks run at 50% each.
- **Deadlines**: success before deadline = reward + prestige. Failure = prestige penalty, no reward.
- **Trust**: completing tasks for a client builds trust -> less work per task, access to gated tasks. Working for one client erodes trust with others.
- **Not all clients are reliable.** Check `client history` for failure patterns.
- **Payroll**: deducted monthly. Funds < 0 = bankruptcy.
- Prestige grows per domain. Higher prestige unlocks better-paying tasks.
\end{lstlisting}

\section{Main Results \& Tool-Usage Statistics}
\label{app:main_result_stats}

\begin{figure}[H]
    \centering
    \includegraphics[width=\textwidth]{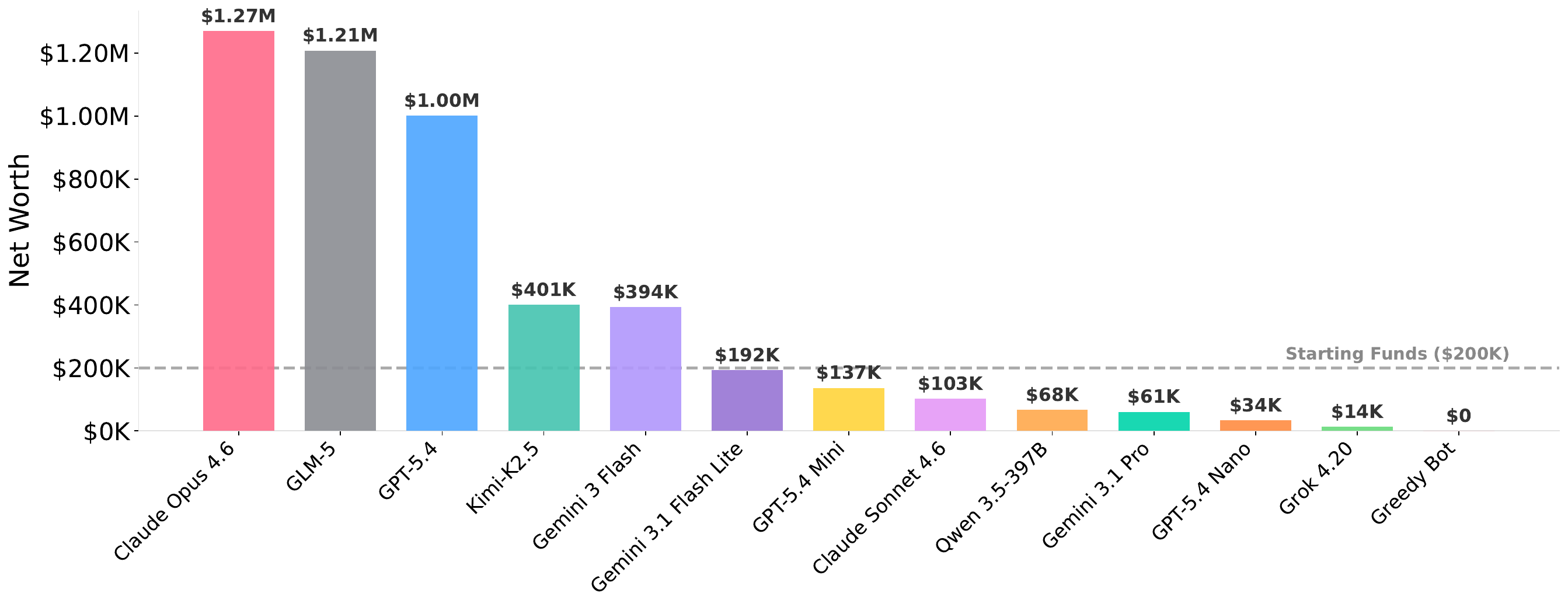}
    \caption{
Average final funds across three seeds for all 12 models and the 
  greedy baseline, sorted by performance. Only three models (Claude Opus 4.6, GLM-5, GPT-5.4)         
  consistently surpass the starting capital.
  }
      \label{fig:leaderboard}
\end{figure}

\begin{figure}[H]
    \centering
    \includegraphics[width=\textwidth]{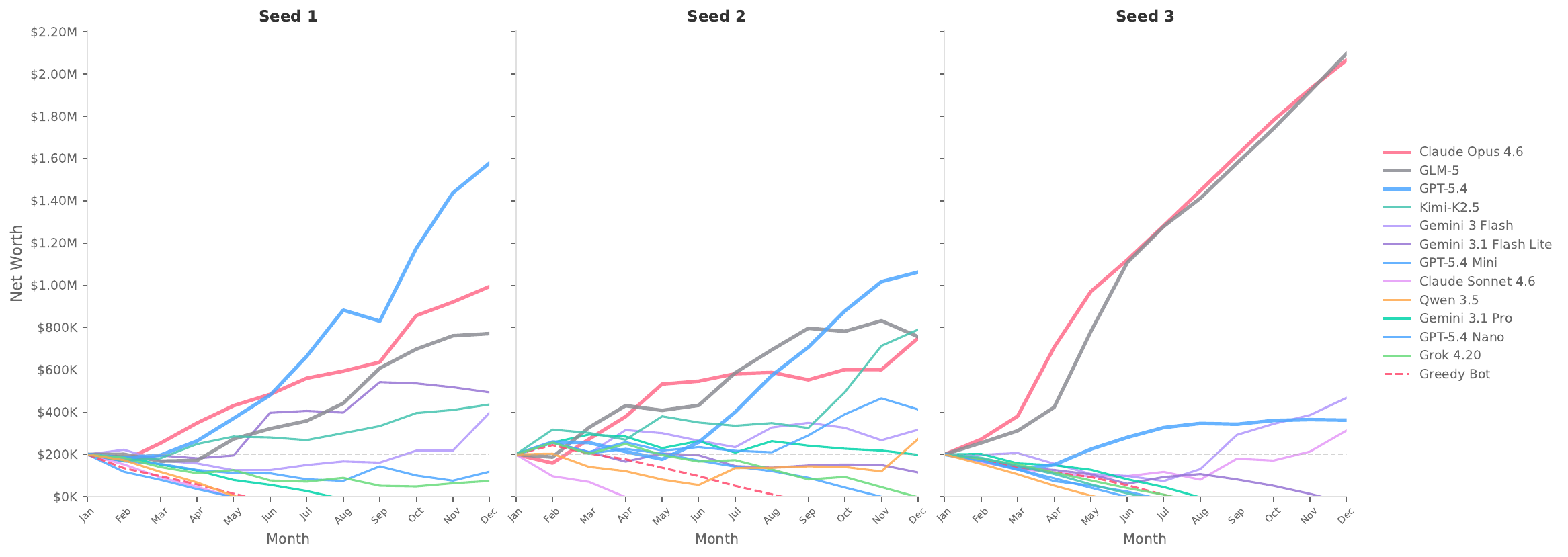}
    \caption{
Monthly funds trajectory across three seeds. 
  }
      \label{fig:progression_per_seed}
\end{figure}

\begin{figure}[H]
    \centering
    \includegraphics[width=\textwidth]{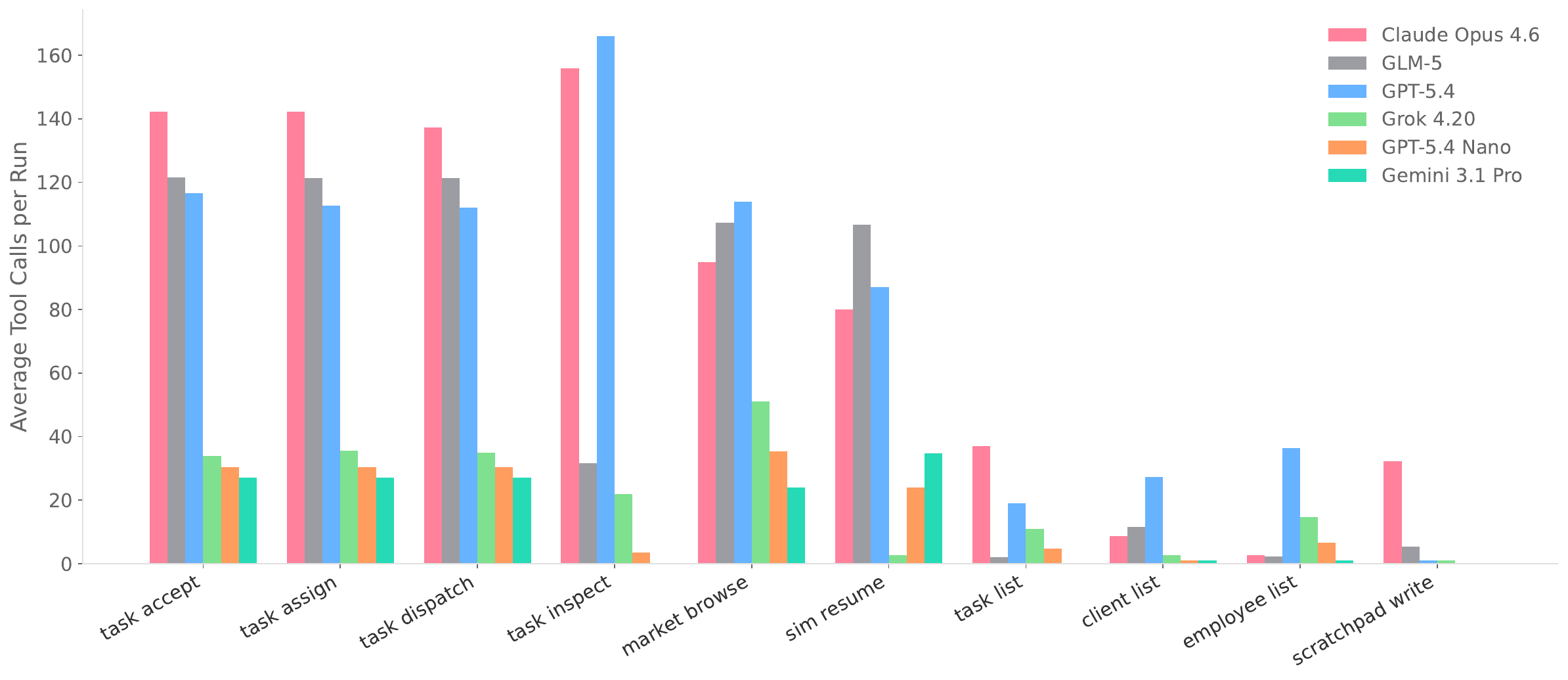}
    \caption{
Average number of tool calls per run for the top three and bottom 
  three performing models, grouped by command type. Top models execute 
  3-4x more actions overall, with the largest gaps in task inspection and 
  scratchpad usage.
  }
    \label{fig:tool_use}
\end{figure}

Figure~\ref{fig:leaderboard} shows the average revenue of each model across all three seeds, with the horizontal line showing the starting fund. 
Figure~\ref{fig:progression_per_seed} shows the revenue progression of all models per seed.
The relative performance difference across models is consistent across seeds.
Figure~\ref{fig:tool_use} and Table~\ref{tab:behavioral_stats} further break down specific behavioral patterns by models.

\begin{table}
\centering
\small
\begin{tabularx}{\textwidth}{@{}lrrrr@{}}
\toprule
\textbf{Model} & \textbf{SP/100T} & \textbf{Inspect/Accept} & \textbf{Avg Conc.} & \textbf{Cmd/Turn} \\
\midrule
Claude Opus 4.6        & 5.6 & 1.10 & 2.0 & 1.41 \\
GLM-5                  & 2.7 & 0.26 & 1.5 & 2.16 \\
GPT-5.4                & 10.6 & 1.43 & 2.5 & 2.63 \\
Kimi-K2.5              & 0.6 & 0.00 & 1.3 & 1.18 \\
Gemini 3 Flash         & 0.2 & 0.11 & 1.2 & 0.24 \\
Gemini 3.1 Flash Lite  & 0.9 & 0.02 & 1.2 & 2.16 \\
GPT-5.4 Mini           & 0.2 & 0.22 & 1.0 & 0.04 \\
Claude Sonnet 4.6      & 4.6 & 0.64 & 7.2 & 2.01 \\
Qwen 3.5-397B          & 0.7 & 0.01 & 2.5 & 1.00 \\
Gemini 3.1 Pro         & 0.0 & 0.00 & 1.1 & 1.06 \\
GPT-5.4 Nano           & 0.0 & 0.08 & 1.2 & 0.08 \\
Grok 4.20 Beta         & 0.4 & 0.65 & 1.5 & 0.92 \\
\bottomrule
\end{tabularx}
\caption{Behavioral statistics per model, averaged across 3 seeds and sorted by average final funds. The top three models use the scratchpad and task inspection significantly more than others. Sonnet's average concurrency of 7.2 is a clear outlier, consistent with its high understaffing failure rate. GPT-5.4 Mini and Nano execute fewer than 0.1 commands per turn despite running thousands of turns, indicating most turns produce reasoning without action.
\emph{SP/100T}: scratchpad writes per 100 turns. \emph{Inspect/Accept}: ratio of \texttt{task inspect} to \texttt{task accept} calls. \emph{Avg Conc.}: mean simultaneously active tasks. \emph{Cmd/Turn}: commands executed per turn.}
\label{tab:behavioral_stats}
\end{table}

\section{Ablation: Context Window Size}
\label{app:context_ablation}

\begin{figure}[h]
  \centering
  \includegraphics[width=0.6\textwidth]{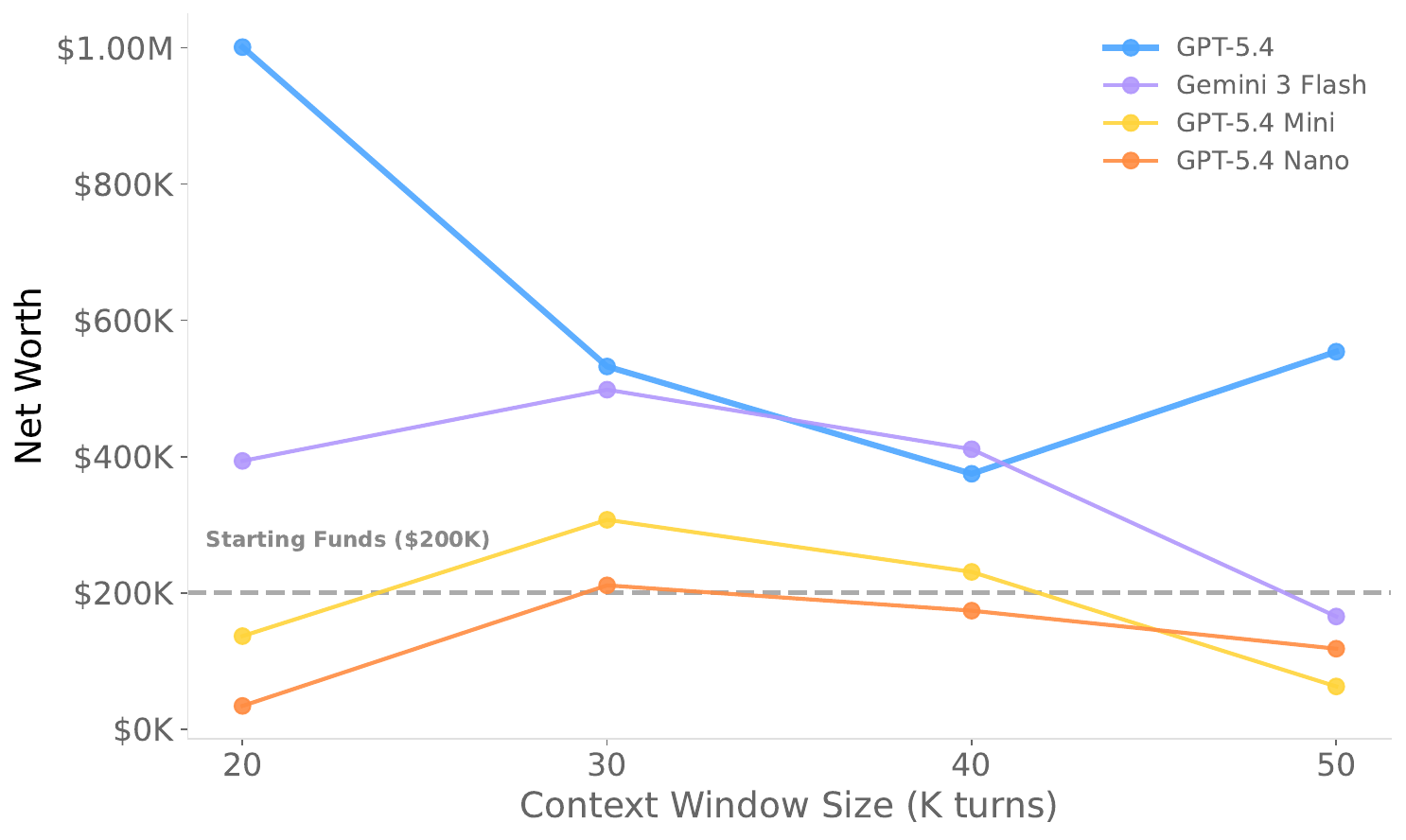}
  \caption{%
    Average final funds vs.\ $K$ (number of retained turns), across three seeds.
  }
  \label{fig:context_ablation}
\end{figure}

We vary the number of retained conversation turns $K \in \{20, 30, 40,
  50\}$ across four models and three seeds each
  (Figure~\ref{fig:context_ablation}).
Three of four models peak at $K = 30$, with GPT-5.4 Mini improving by
  123\% and Gemini 3 Flash by 27\%, before degrading sharply at higher
  values.
  GPT-5.4 performs best at the default $K = 20$: at $K = 30$, it issues
  fewer commands per turn and never learns to bundle
  accept-assign-dispatch into a single action, reducing task throughput
  by over 60\%.
  All four models degrade beyond their optimal $K$, suggesting that
  larger context windows reduce per-turn action efficiency faster than
  they improve decision quality.
  
\section{Summary of Token Usage, Cost, and Time}
\label{app:summary_token_use}

Table~\ref{tab:cost} summarizes the API cost, number of total tokens, and runtime in minutes of each model.

\begin{table}[H]
\centering
\begin{tabular}{lcrrr}
\toprule
\textbf{Model} & \textbf{Bankrupt} & \textbf{Cost (\$)} & \textbf{Tokens} & \textbf{Runtime (min)} \\
\midrule
Claude Opus 4.6       & 0/3 & 86.07 $\pm$ 41.17 & 16.7M & 70 $\pm$ 36 \\
GLM-5                 & 0/3 &  7.62 $\pm$  1.06 &  8.3M & 40 $\pm$  7 \\
GPT-5.4               & 0/3 & 23.08 $\pm$  7.07 &  9.5M & 19 $\pm$  5 \\
Kimi-K2.5             & 1/3 &  1.79 $\pm$  2.74 &  3.8M & 21 $\pm$ 30 \\
Gemini 3 Flash        & 0/3 &  1.83 $\pm$  0.32 &  3.8M &  6 $\pm$  1 \\
Gemini 3.1 Flash Lite & 1/3 &  0.59 $\pm$  0.22 &  2.6M &  6 $\pm$  2 \\
GPT-5.4 Mini          & 2/3 &  1.34 $\pm$  1.83 &  1.8M &  3 $\pm$  4 \\
Claude Sonnet 4.6     & 2/3 & 18.06 $\pm$ 18.87 &  5.8M & 23 $\pm$ 24 \\
Qwen 3.5-397B         & 2/3 &  1.16 $\pm$  0.79 &  2.8M & 13 $\pm$  8 \\
Gemini 3.1 Pro        & 2/3 &  7.36 $\pm$  8.18 &  2.8M &  7 $\pm$  6 \\
GPT-5.4 Nano          & 2/3 &  0.41 $\pm$  0.20 &  2.0M &  3 $\pm$  1 \\
Grok 4.20             & 2/3 &  5.03 $\pm$  2.40 &  2.9M &  4 $\pm$  2 \\
\bottomrule
\end{tabular}
\caption{Computational cost per run, averaged across three seeds ($\mu \pm \sigma$). Models are sorted by average final funds.}
\label{tab:cost}
\end{table}

\end{document}